\documentclass[lettersize,journal]{IEEEtran}
\usepackage{amsmath,amsfonts}
\usepackage{algorithmic}
\usepackage{algorithm}
\usepackage{array}
\usepackage[caption=false,font=normalsize,labelfont=sf,textfont=sf]{subfig}
\usepackage{textcomp}
\usepackage{stfloats}
\usepackage{url}
\usepackage{verbatim}
\usepackage{graphicx}
\usepackage{cite}
\usepackage{subcaption}
\usepackage{amsmath,amsfonts,bm}

\newcommand{\bR}{\mathbb{R}}

\newcommand{\vA}{{\bf A}}
\newcommand{\vB}{{\bf B}}
\newcommand{\vC}{{\bf C}}

\newcommand{\vI}{{\bf I}}

\newcommand{\vX}{{\bf X}}
\newcommand{\vY}{{\bf Y}}

\def\eqref#1{equation~\ref{#1}}

\def\1{\bm{1}}

\def\vh{{\bm{h}}}

\def\vx{{\bm{x}}}

\DeclareMathAlphabet{\mathsfit}{\encodingdefault}{\sfdefault}{m}{sl}
\SetMathAlphabet{\mathsfit}{bold}{\encodingdefault}{\sfdefault}{bx}{n}

\usepackage{multirow}
\usepackage[pagebackref=true,breaklinks=true,colorlinks,bookmarks=false]{hyperref}
\usepackage{ulem}

\usepackage{pifont}

\newcommand{\ie}{i.e.}
\newcommand{\convblock}[4]{
\multirow{3}{*}{
$
\left[
\begin{array}{c}
1\times 1, (C_{out},C_{in}) = #1,64 \\
[-.1em] 3\times 3, (C_{out},C_{in}) = #1,#1 \\
[-.1em] 1\times 1, (C_{out},C_{in}) = 4\times#2,#2
\end{array}
\right]
\times #3\times #4
$
}}
\usepackage{makecell}
\newcommand{\HRVMambablock}[6]{
\setlength{\tabcolsep}{2pt}
\multirow{3}{*}{
$
\left[
\begin{array}{c}
\text{ESInB}\\
\text{DSS2D}, #5\\
\text{FFN}, #4
\end{array}
\right]
\times#2\times #3
$
}
}

\begin{document}
\title{Efficient High-Resolution Visual Representation Learning with State Space Model for Human Pose Estimation}

\author{
    Hao Zhang,
    ~
    Yongqiang Ma,
    ~
    Wenqi Shao,
    ~
    Ping Luo,
    ~
    Nanning Zheng,~\IEEEmembership{Fellow,~IEEE,}
    ~
    Kaipeng Zhang  
\thanks{
Corresponding authors: Kaipeng Zhang; Nanning Zheng. 

Hao Zhang, Yongqiang Ma, and Nanning Zheng are with National Key Laboratory of Human-Machine Hybrid Augmented Intelligence, National Engineering Research Center for Visual Information and Applications, and Institute of Artificial Intelligence and Robotics, Xi'an Jiaotong University, Xi’an, Shaanxi 710049, China (e-mail: zhanghao520@stu.xjtu.edu.cn, musayq@xjtu.edu.cn, nnzheng@mail.xjtu.edu.cn).

Wenqi Shao, Ping Luo, Kaipeng Zhang are with Shanghai Artificial Intelligence Laboratory, Shanghai, 200000, China (e-mail: shaowenqi@pjlab.orn.cn, pluo@cs.hku.edu, zhangkaipeng@pjlab.org.cn).
}
}

\maketitle
\begin{abstract}
Capturing long-range dependencies while preserving high-resolution visual representations is crucial for dense prediction tasks such as human pose estimation. Vision Transformers (ViTs) have advanced global modeling through self-attention but suffer from quadratic computational complexity with respect to token count, limiting their efficiency and scalability to high-resolution inputs, especially on mobile and resource-constrained devices.
State Space Models (SSMs), exemplified by Mamba, offer an efficient alternative by combining global receptive fields with linear computational complexity, enabling scalable and resource-friendly sequence modeling. However, when applied to dense prediction tasks, existing visual SSMs face key limitations: weak spatial inductive bias, long-range forgetting from hidden state decay, and low-resolution outputs that hinder fine-grained localization.
To address these issues, we propose the \textit{Dynamic Visual State Space} (DVSS) block, which augments visual state space models with multi-scale convolutional operations to enhance local spatial representations and strengthen spatial inductive biases. Through architectural exploration and theoretical analysis, we incorporate deformable operation into the DVSS block, identifying it as an efficient and effective mechanism to enhance semantic aggregation and mitigate long-range forgetting via input-dependent, adaptive spatial sampling.
We embed DVSS into a multi-branch high-resolution architecture to build \textbf{HRVMamba}, a novel model for efficient high-resolution representation learning. Extensive experiments on human pose estimation, image classification, and semantic segmentation show that HRVMamba performs competitively against leading CNN-, ViT-, and SSM-based baselines.
The code is available at \url{https://github.com/zhanghao5201/PoseVMamba}.
\end{abstract}


\section{Introduction}
\label{sec:intro}

\begin{figure}[t]
\centering
\includegraphics[width=0.42\textwidth]{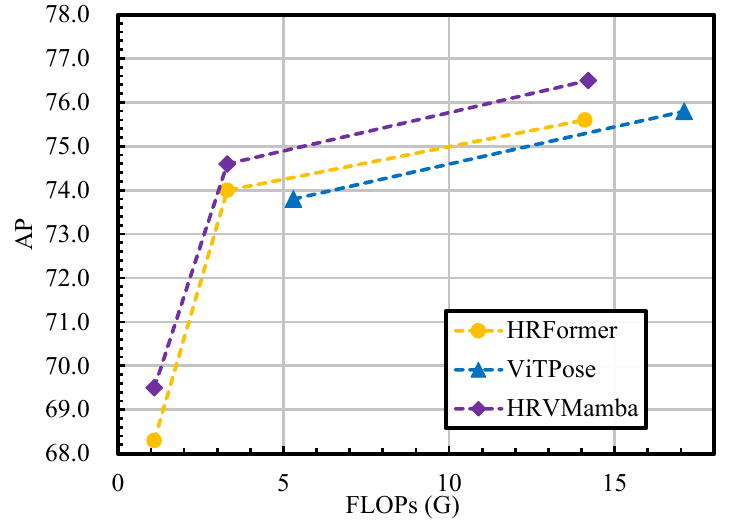}
\includegraphics[width=0.42\textwidth]{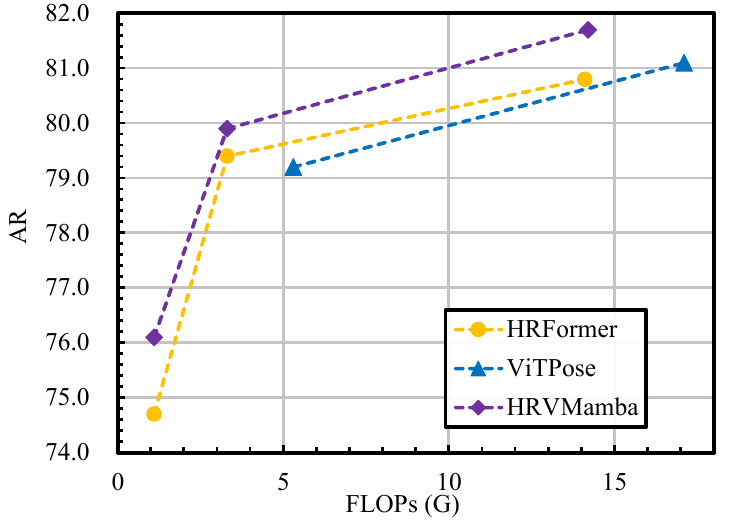}
\caption{\textbf{The trade-off between AP, AR and FLOPs on COCO pose estimation \texttt{val} set for human pose estimation.}
}\
\label{ap_ar}
\end{figure}
Learning robust high-resolution representations is a fundamental yet challenging requirement for dense prediction tasks such as human pose estimation~\cite{MVPose,xu024vitpose+,zhang2024open}. Conventional visual backbones like Convolutional Neural Networks (CNNs)~\cite{he2016deep,zhang2023scgnet,zhang2024hf,zhang2024fmgnet,DBLP:journals/tmc/RohBKK25} are effective at capturing local patterns, thanks to their strong spatial inductive biases and linear computational complexity. However, their limited receptive fields restrict their ability to model long-range dependencies, which are critical for complex visual understanding in dense prediction tasks. 
On the other hand, as illustrated in Fig.~\ref{ap_ar}, Vision Transformers (ViTs)\cite{yuan2021hrformer,liu2021Swin,DBLP:journals/tmc/WangZWDZ24,DBLP:journals/tmc/GongLLDWJ25} exploit global self-attention to model long-range context. However, their quadratic complexity with respect to token count makes them computationally expensive, which in turn limits their performance, especially in high-resolution scenarios and when deployed on resource-constrained platforms such as mobile devices.

State Space Models (SSMs) have recently emerged as efficient alternatives to ViTs, offering linear computational complexity with respect to token length while modeling long-range global dependencies. By combining linear computational complexity with global receptive capabilities, SSMs enable scalable and computation-friendly modeling of long sequences. This advancement has led to a surge of visual SSM architectures, such as ViM~\cite{zhu2024vision}, VMamba~\cite{liu2024vmamba}, LocalVMamba~\cite{huang2024localmamba}, VideoMamba~\cite{li2024videomamba}, GroupMamba~\cite{shaker2025groupmamba}, and MambaVision~\cite{hatamizadeh2025mambavision}.

\begin{figure*}[t]
    \centering
    \includegraphics[width=0.9\textwidth]{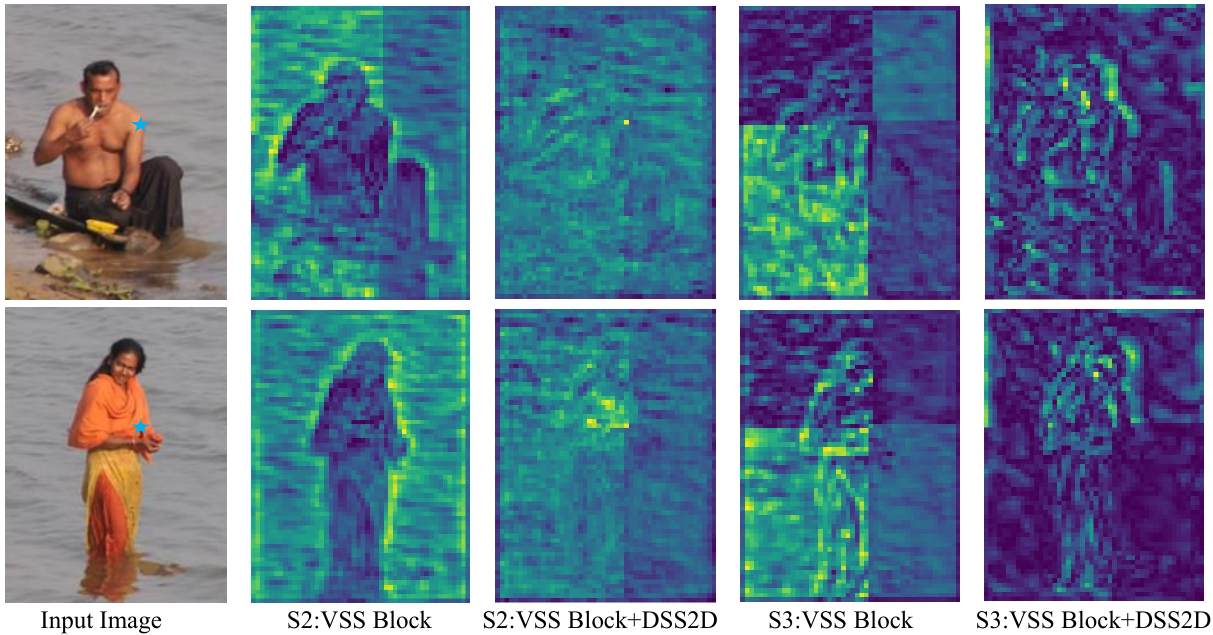}\\ 
    \caption{
    \textbf{Activation maps of SSM for the query location} (marked by a blue pentagram). We embed the VSS block~\cite{liu2024vmamba} and VSS+DSS2D (where the SS2D block is replaced by our proposed Deformable 2D-Selective-Scan (DSS2D) block) into HRVMamba-Small. Here, $S_i$ denotes the $i$-th stage of HRVMamba. We visualize the SSM activation maps from the second block of the first block in the first branch of each $S_i$. In the early stage (S2), the DSS2D block attends to semantically meaningful regions that are relevant to the query patch, while the SS2D block primarily focuses on low-level edge features. In the later stage (S3), the DSS2D block effectively emphasizes human-related regions, whereas the SS2D block still responds to less informative background areas.
    }
    \label{fig_coco}
\end{figure*}

Nevertheless, despite their improved efficiency, existing SSM-based visual models face several key limitations in dense prediction tasks such as human pose estimation. First, their tokenized sequence representation and bi-/multi-directional scanning mechanisms~\cite{zhu2024vision,huang2024localmamba} disrupt the spatial continuity of images, resulting in insufficient spatial inductive bias required for capturing local image details. Secondly, the recurrent nature of token processing causes gradual information decay in hidden states, leading to the phenomenon of long-range forgetting. Consequently, these models may lose crucial high-level semantic information relevant to distant tokens, defaulting instead to low-level features such as edges. This effect is visually demonstrated in Fig.~\ref{fig_coco}, column 2\footnote{We visualize attention maps using VMamba's approach~\cite{liu2024vmamba}.}. Lastly, existing visual SSMs often produce low-resolution, single-scale feature maps, which inadequately represent fine-grained spatial details and multi-scale variability necessary for precise pose estimation.

To address these issues, we propose the \textit{Dynamic Visual State Space} (DVSS) block, an improved variant of the Visual State Space (VSS) block in VMamba. DVSS incorporates convolutional kernels at multiple scales to robustly capture local spatial features and strengthen the model’s inductive bias across different resolutions. 
To further alleviate long-range forgetting and enhance spatial dependency modeling, we systematically explore a range of architectural design alternatives. Through empirical comparisons and theoretical insights, we identify deformable operation as a particularly effective mechanism. 
It dynamically adjusts spatial aggregation based on the input and task-specific context, enabling adaptive sampling and enhancement of semantically relevant yet spatially distant regions.
Our analysis demonstrates how this mechanism mitigates the decay of contextual information over distance and allows the model to focus selectively on meaningful spatial cues, thereby enhancing both representation robustness and performance.  
For example, as shown in Fig.~\ref{fig_coco}, the left shoulder and chest features near the right shoulder are highlighted (row 1, column 3), the head features connected to the right shoulder are emphasized (row 1, column 5), along with the highlighted features of both hands and the chest (row 2, column 3). 

Building on the multi-resolution parallel framework, we embed DVSS blocks into multi-branch architectures to form the High-Resolution Visual State Space Model named \textbf{HRVMamba}. This model effectively preserves high-resolution information and models multi-scale feature variations, thus excelling in dense visual tasks. HRVMamba achieves a better trade-off between computational complexity and accuracy, making it suitable for scenarios with limited computational budgets (e.g., under similar FLOP constraints), while still maintaining detailed spatial modeling.

The contributions of this study are as follows: 
\begin{itemize} 
\item We propose \textbf{HRVMamba}, a high-resolution visual state space model that efficiently supports high-resolution representation learning. It adopts a multi-resolution branch architecture to preserve fine-grained details and capture multi-scale variations for human pose estimation.
\item We introduce the DVSS block, which combines multi-scale convolutional kernels and deformable operations to enhance inductive bias and mitigate the long-range forgetting problem. 
\item HRVMamba demonstrates promising performance in human pose estimation, image classification, and semantic segmentation tasks. Experimental results show that HRVMamba achieves competitive results against existing CNN, ViT, and SSM benchmark models.
\end{itemize}

The remainder of this paper is organized as follows: Section~\ref{related} presents a concise overview of related work in CNNs, ViTs, SSMs, and high-resolution representation learning. Section~\ref{Preliminaries} introduces the theoretical foundations of SSMs and the Selective State Space Model. Section~\ref{method} details the proposed HRVMamba architecture, including its core component, \ie, the DVSS Block. Section~\ref{experiments} reports extensive experimental results and ablation studies across multiple benchmarks. Finally, Section~\ref{conclusion} concludes the paper.

\section{Related work}
\label{related}
In this section, we first review related work on Convolutional Neural Networks and Vision Transformers.
Next, we discuss recent advances in State Space Models.
Finally, we introduce works specifically focused on high-resolution representation learning.

\subsection{Convolutional Neural Networks (CNNs) and Vision Transformers (ViTs)}
CNNs have long been the cornerstone of computer vision, evolving from early models like AlexNet~\cite{xiong2024efficient} and ResNet~\cite{he2016deep} to more advanced architectures such as ConvNeXt~\cite{liu2022convnet}, SCGNet~\cite{zhang2023scgnet}, FlashInternImage~\cite{xiong2024efficient}, and FMGNet~\cite{zhang2024fmgnet}. These models excel at local feature extraction, leveraging strong inductive biases and efficient computation, and have achieved remarkable performance across a wide range of tasks including image classification, semantic segmentation, and human pose estimation.
ViTs introduce self-attention mechanisms from natural language processing, segmenting images into non-overlapping patches to effectively model long-range global dependencies. This paradigm shift forms the backbone of modern Large Vision-Language Models~\cite{zhang2024avibench,ying2024mmt,liu2024convbench}. Numerous enhancements, such as the distillation strategies in DeiT~\cite{touvron2021training}, the hierarchical design in Swin Transformer~\cite{liu2021Swin}, and the lightweight attention mechanisms in SwiftFormer~\cite{shaker2023swiftformer}, have been proposed to improve both performance and efficiency, promoting broader adoption of ViTs in various vision tasks.
Meanwhile, hybrid architectures~\cite{yun2024shvit, MaDY0C0Y24} that combine the complementary strengths of CNNs and ViTs have gained increasing attention. These models leverage the local feature extraction efficiency and inductive bias of CNNs while incorporating the global context modeling capabilities of ViTs, marking a promising and significant direction in the development of modern backbone networks.

\subsection{State Space Models (SSMs)}
SSMs provide a mathematical framework for modeling dynamic systems and exhibit linear computational complexity with respect to sequence length, making them highly efficient for processing long sequential data. Advances in models such as S4~\cite{gu2021efficiently}, S5~\cite{smith2022simplified}, and H3~\cite{fu2022hungry} have significantly improved SSMs by incorporating structure-aware optimizations, parallel scan mechanisms, and hardware-friendly designs that enhance both accuracy and speed.
Mamba~\cite{gu2023mamba} further advances the field by introducing input-dependent parameterization and a hardware-efficient parallel scanning mechanism known as S6, firmly establishing SSMs as a compelling and scalable alternative to Transformer-based architectures. Since Mamba’s introduction, SSMs have been increasingly adopted in visual domains, with early efforts such as S4ND~\cite{nguyen2022s4nd} treating images as continuous 2D signals and demonstrating the viability of SSMs for vision tasks.
Building upon Mamba’s success, several vision-oriented variants have been proposed. ViM~\cite{zhu2024vision} and VMamba~\cite{liu2024vmamba} mitigate the directional bias of unidirectional scanning by incorporating bidirectional or four-way scan strategies. LocalVMamba~\cite{huang2024localmamba} introduces localized window-based scanning to better preserve fine-grained spatial details, while PlainMamba~\cite{yang2024plainmamba} reworks 2D scanning for sequential data processing in a simpler form. 
GroupMamba~\cite{shaker2025groupmamba} enhances training stability and convergence through a distillation-based optimization framework, and MambaVision~\cite{hatamizadeh2025mambavision} combines the strengths of SSMs and Transformers for hybrid feature modeling.
Despite these innovations, most existing visual Mamba-based models still suffer from long-range forgetting and rely primarily on single-scale, low-resolution feature representations. This significantly limits their ability to preserve fine-grained visual cues and to capture the multi-scale variations that are crucial for dense prediction tasks such as human pose estimation.

\subsection{High-Resolution Representation Learning} 
Learning effective high-resolution representations is essential for dense prediction tasks such as human pose estimation~\cite{MVPose,xu024vitpose+,zhang2024open}.
The High-Resolution Network (HRNet)~\cite{wang2020deep} is the first to propose maintaining high-resolution representations throughout the entire network, demonstrating impressive performance in dense prediction tasks such as human pose estimation and semantic segmentation. HRNet employs a multi-resolution parallel architecture, where information from different scales is continuously fused through repeated multi-scale feature exchange, allowing it to effectively capture both fine-grained details and global context.
Building on this foundation, HRFormer~\cite{yuan2021hrformer} incorporates the self-attention mechanism~\cite{liu2021Swin} into the high-resolution architecture, combining the benefits of Transformers with HRNet’s strong structural inductive biases. This design further enhances performance across various dense prediction tasks by enabling better long-range context modeling.
Subsequent efforts have focused on improving the efficiency and deployability of high-resolution architectures. Lite-HRNet~\cite{yu2021lite}, Dite-HRNet~\cite{LiZ0ZB22dite}, and HF-HRNet~\cite{zhang2024hf} introduce lightweight convolutional backbones using techniques such as depthwise convolution and dynamic convolution. 
Despite these advances, it remains unclear whether Mamba can fully exploit high-resolution structures for efficient high-resolution representation learning. Key challenges include addressing Mamba's inherent lack of spatial inductive bias and its tendency toward long-range forgetting, particularly in complex visual scenes. Effectively integrating Mamba into multi-resolution frameworks while preserving both fine-grained detail and high-level semantics remains an open and promising research direction.

\begin{figure*}[t]
    \centering
    \includegraphics[width=0.92\textwidth]{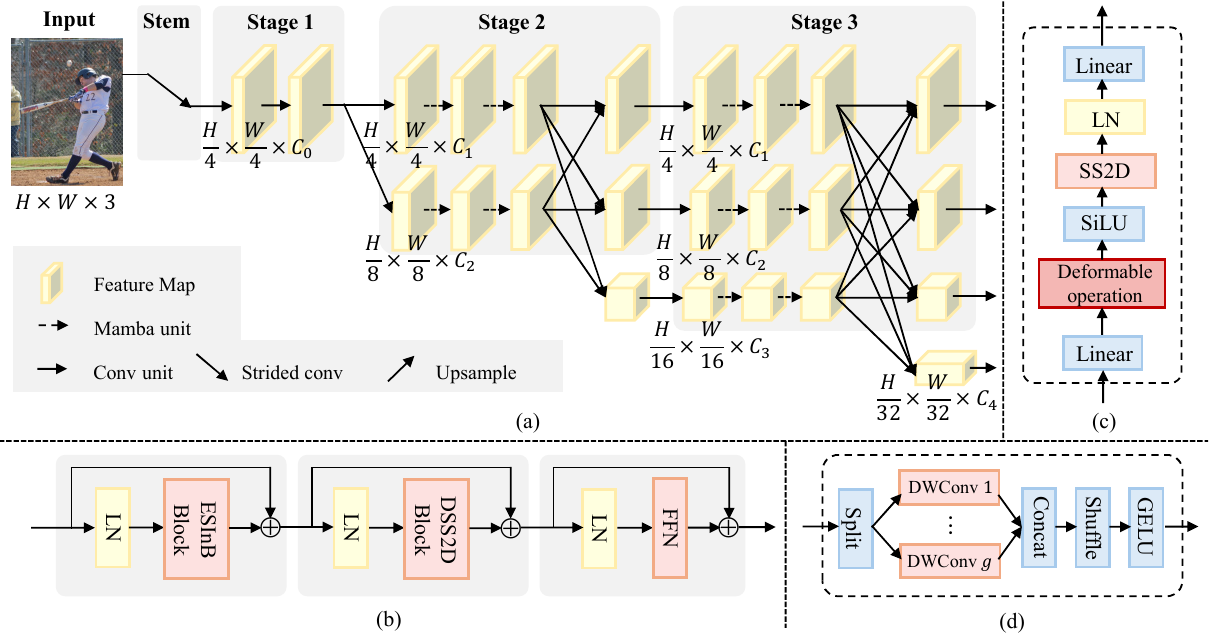}
    \caption{\textbf{(a) Overall architecture of HRVMamba.} 
    \textbf{(b) Dynamic Visual State Space block.} 
    \textbf{(c) Deformable 2D-Selective-Scan (DSS2D) Block.}
    \textbf{(d) Enhanced Spatial Inductive Bias Block (ESInB) Block.}
    HRVMamba has four stages, but for demonstration purposes, we only show three. $H$ and $W$ represent the height and width of the image, while $C_i$ denotes the number of channels in the $i$-th branch or position.
    LN, Linear, DWConv and SS2D represent LayerNorm, Linear Layer, depthwise convolution, and 2D-Selective-Scan SSM.
    }
    \label{fig:model}
\end{figure*}

\section{Preliminaries}
\label{Preliminaries}
\textbf{State Space Models (SSMs)} map input stimulation $x\in \bR^{1}$ to output response $y\in \bR^{1}$ through a hidden state $\vh\in\bR^{N\times 1}$ based on continuous linear time-invariant (LTI) systems, where $N$ represents the number of states. 
To integrate deep models and adapt to real-world data, discretization must be applied to convert the continuous differential equations of SSMs into discrete functions using the zero-order hold method. Specifically, with a discrete-time step $\Delta \in \bR^{1}$, SSMs are discretized as follows:
\begin{align}
\vh(t) &= \tilde{\vA}\vh(t-1) + \tilde{\vB} \vX(t), \label{eq:ssm-d1}  \\ 
\vY(t) &= \vC^\top \vh(t), 
\label{eq:ssm-d2}
\end{align}
where $\vX(t) = x(\Delta t)$, $\vA \in \bR^{N\times N}$ is the system’s evolution matrix, and $\vB \in \bR^{N\times 1}$ and $\vC \in \bR^{N\times 1}$ are the projection matrices.
$\tilde{\vA} = \exp(\Delta \vA), \quad \tilde{\vB} = (\Delta \vA)^{-1}(\exp(\Delta \vA) - \vI) \cdot \Delta \vB \approx \Delta \vB$, where $\vI$ denotes the identity matrix.

\textbf{Selective State Space Models (S6)} are introduced in Mamba~\cite{gu2023mamba} to improve the extraction of strong contextual information. S6 allows $\vB$, $\vC$, and $\Delta$ to vary as functions of the input $\vX(t)$, whereas in S4~\cite{gu2021efficiently}, $\vA$, $\vB$, $\vC$, and $\Delta$ are input-independent, which limits the model's ability to extract crucial information from the input sequence.
Formally, given an input sequence $\vX \in \bR^{B \times L \times C}$, where $B$, $L$, and $C$ represent the batch size, sequence length, and feature dimension, respectively, the input-dependent parameters $\vB$, $\vC$, and $\Delta$ are computed as follows:
\begin{align} 
\vB &= \texttt{Linear}(\vX) \in \bR^{B \times L \times N}, \\ 
\vC &= \texttt{Linear}(\vX) \in \bR^{B \times L \times N}, \\
\Delta &= \texttt{SoftPlus}(\tilde{\Delta} + \texttt{Linear}(\vX)) \in \bR^{B \times L \times C}, 
\end{align}
where $\tilde{\Delta} \in \bR^{B \times L \times C}$ is a learnable parameter, and $\vA \in \bR^{C\times N}$ is the the system’s evolution matrix.

\begin{table*}[t]
  \setlength{\tabcolsep}{0.2mm}
  \renewcommand\arraystretch{1.1}
  \caption{\textbf{The architecture configuration of HRVMamba.}
    ESInB, and DSS2D represent the ESInB block and DSS2D block respectively. 
    $\left(M_1, M_2, M_3, M_4\right)$: the number of blocks,
    $\left(B_1, B_2, B_3, B_4\right)$: the number of blocks,
    $\left(S_1, S_2, S_3, S_4\right)$: the SSM expansion ratios,
    $\left(R_1, R_2, R_3, R_4\right)$: the MLP expansion ratios.    
  }
  \label{HRVMamba_arch}
  \centering
  {
  {
    \begin{tabular}{l|c|c|c|c}
      \hline
      {Res.}                                                             & {Stage $1$}                               & Stage $2$                                                           & Stage $3$                                                           & Stage $4$                                                           \\
      \hline
      \multirow{3}{*}{$4\times$}
      & \convblock{$$C_{0}$$}{$$C_{0}$$}{$$B_1$$}{$$M_1$$} &
      \HRVMambablock{$$C$$}{$$B_2$$}{$$M_2$$}{$$R_1$$}{$$S_1$$}{$$G_1$$}   &
      \HRVMambablock{$$C$$}{$$B_3$$}{$$M_3$$}{$$R_1$$}{$$S_1$$}{$$G_1$$}   &
      \HRVMambablock{$$C$$}{$$B_4$$}{$$M_4$$}{$$R_1$$}{$$S_1$$}{$$G_1$$}          \\
      &                                           &  &  &  \\
      &                                           &  &  &  \\
      \hline
      \multirow{3}{*}{$8\times$}
      &                                           & \HRVMambablock{$$2C$$}{$$B_2$$}{$$M_2$$}{$$R_2$$}{$$S_2$$}{$$G_2$$} & \HRVMambablock{$$2C$$}{$$B_3$$}{$$M_3$$}{$$R_2$$}{$$S_2$$}{$$G_2$$} & \HRVMambablock{$$2C$$}{$$B_4$$}{$$M_4$$}{$$R_2$$}{$$S_2$$}{$$G_2$$} \\
      &                                           &  &  &  \\
      &                                           &  &  &  \\
      \hline
      \multirow{3}{*}{$16\times$}
      &                                           &  & \HRVMambablock{$4C$}{$$B_3$$}{$$M_3$$}{$$R_3$$}{$$S_3$$}{$$G_3$$}   & \HRVMambablock{$4C$}{$$B_4$$}{$$M_4$$}{$$R_3$$}{$$S_3$$}{$$G_3$$}   \\
      &                                           &  &  &  \\
      &                                           &  &  &  \\
      \hline
      \multirow{3}{*}{$32\times$}
      &                                           &  &  & \HRVMambablock{$$8C$$}{$$B_4$$}{$$M_4$$}{$$R_4$$}{$$S_4$$}{$$G_4$$} \\
      &                                           &  &  &  \\
      &                                           &  &  &  \\
      \hline
    \end{tabular}
  }}
\end{table*}
\begin{table*}[t]
  \centering
  \caption{\textbf{Architecture details of HRVMamba variants.} $\left(C_0,C_1, C_2, C_3, C_4\right)$ is defined in Section~\ref{method_1}. 
  }
  \label{HRVMamba_variants}
  \renewcommand\arraystretch{1.1}
  \setlength{\tabcolsep}{3.65mm}
  \begin{tabular}{l|c|c|c|c|c}
    \hline
    Model     & \thead{\#channels \\ $\left(C_0,C_1, C_2, C_3, C_4\right)$}  & \thead{\#blocks\\$\left(B_1, B_2, B_3, B_4\right)$} & \thead{\#blocks   \\ $\left(M_1, M_2, M_3, M_4\right)$} & \thead{\#SSM ratio\\$\left(S_1, S_2, S_3, S_4\right)$} & \thead{\#MLP ratio\\$\left(R_1, R_2, R_3, R_4\right)$}\\
    \hline
     HRVMamba-Nano & $\left(16,8,16,32, 64\right)$ & $\left(2,2,2,2\right)$   & $\left(1,1,4,2\right)$ & $\left(2,2,2,2\right)$ & $\left(2,2,2,2\right)$ \\
    HRVMamba-Tiny & $\left(32,16,32, 64, 128\right)$ & $\left(2,2,2,2\right)$   & $\left(1,1,4,2\right)$ & $\left(2,2,2,2\right)$ & $\left(2,2,2,2\right)$ \\
    HRVMamba-Small & $\left(64,32,64,128,256\right)$ & $\left(2,2,2,2\right)$   & $\left(1,1,4,2\right)$ & $\left(2,2,2,2\right)$ & $\left(2,2,2,2\right)$ \\
    HRVMamba-Base & $\left(64,80,160,320,640\right)$ & $\left(2,2,2,2\right)$ & $\left(1,1,4,2\right)$   & $\left(2,2,2,2\right)$ & $\left(2,2,2,2\right)$ \\
    \hline
  \end{tabular}
\end{table*}

\section{High-Resolution Visual State Space Model}
\label{method}
\subsection{Multi-resolution Parallel VMamba}
\label{method_1}
Existing visual Mamba models typically produce single-scale, low-resolution feature maps, which leads to significant information loss and hampers their ability to capture the fine-grained details and multi-scale variations required for dense prediction tasks. To overcome this limitation, we incorporate the multi-resolution parallel design from HRNet~\cite{wang2020deep} and propose High-Resolution Visual State Space Model (HRVMamba), which maintains high-resolution representations throughout the network.

The overall architecture of HRVMamba is illustrated in Fig.~\ref{fig:model}(a). Given an input image $\vX \in \bR^{H \times W \times 3}$, the network begins with a downsampling stem consisting of two 3$\times$3 convolutional layers with stride 2, reducing the spatial resolution to $\frac{H}{4} \times \frac{W}{4}$. The backbone is composed of four stages, where each stage progressively introduces a lower-resolution branch while retaining all higher-resolution streams from the previous stage. By the final stage, the network maintains four parallel branches with spatial dimensions of $\frac{H}{4} \times \frac{W}{4} \times C_1$, $\frac{H}{8} \times \frac{W}{8} \times C_2$, $\frac{H}{16} \times \frac{W}{16} \times C_3$, and $\frac{H}{32} \times \frac{W}{32} \times C_4$, respectively.
Motivated by prior work~\cite{yun2024shvit,MaDY0C0Y24}, which highlights the effectiveness of convolutional operations on higher-resolution feature maps in early stages, we employ a standard Bottleneck block (as in HRNet) in the first stage. In the remaining stages, we use our proposed Dynamic Visual State Space (DVSS) block (Fig.~\ref{fig:model}(c)) as the primary building unit. To enable information exchange across resolutions, we follow HRNet’s multi-scale fusion strategy, which leverages a series of upsampling and downsampling blocks to aggregate features from different branches.

\begin{table*}[t]
  \centering
  \caption{\textbf{Comparison on the COCO pose estimation \texttt{val} set.} 
    ''\textit{Trans.}'' means transformer architecture.
    $-$ means the numbers are not provided in the original paper. 
    $^\dag$ marks a model that is not pretrained on ImageNet~\cite{deng2009imagenet}, while 
    $^\ddag$ signifies that the backbone uses the classic decoder from ViTPose~\cite{xu024vitpose+}. 
    The \#param. and FLOPs of HRFormer~\cite{yuan2021hrformer} are based on the implementation from MMPOSE~\cite{mmpose2020}.
  }
  \label{coco_pose_val}
  \setlength{\tabcolsep}{3.2mm}
  {
    \begin{tabular}{c|l|c|cc|cccccc}
      \hline
      Arch. & Method      & Input Size& \#param.  & FLOPs    &
      $\operatorname{AP}$        & $\operatorname{AP}^{50}$ & $\operatorname{AP}^{75}$ & $\operatorname{AP}^{M}$ & $\operatorname{AP}^{L}$ & $\operatorname{AR}$\\
      \hline       
      \multirow{2}{*}{\textit{\textbf{CNN}}}&HRNet-W$48$~\cite{wang2020deep}   & $256\times 192$& $63.6$M   & $14.6$G  & ${75.1}$ & ${90.6}$ & ${82.2}$ & ${71.5}$ & ${81.8}$ & ${80.4}$ \\
      ~&FlashInternImage-B$^\ddag$~\cite{xiong2024efficient} &$256\times 192$& 100.7M &17.0G & 74.1 & 90.6 & 82.0 & 70.3 & 80.4 &79.3  \\     
      \hline
      \multirow{13}{*}{\textit{\textbf{Trans.}}}&PRTR~\cite{li2021pose}     & $512\times 384$& $57.2$M   & $37.8$G  & ${73.3}$ & ${89.2}$ & ${79.9}$ & ${69.0}$ & ${80.9}$ & ${80.2}$ \\
      ~&TransPose-H-A$6$~\cite{yang2021transpose} & $256\times 192$& $17.5$M   & $21.8$G  & ${75.8}$ & ${-}$& ${-}$    & ${-}$    & ${-}$    & ${80.8}$ \\
      ~&TokenPose-L/D$24$~\cite{li2021tokenpose}  & $256\times 192$& $27.5$M   & $11.0$G  & ${75.8}$ & ${90.3}$ & ${82.5}$ & ${72.3}$ & ${82.7}$ & ${80.9}$ \\
      ~&HRFormer-Tiny~\cite{yuan2021hrformer}  & $256\times 192$& 2.1M   & 1.1G  &  68.3 & 87.9 &76.0  & 65.0 & 74.7 & 74.7 \\
      ~&HRFormer-Small~\cite{yuan2021hrformer}  & $256\times 192$& $7.7$M    & $3.3$G   &${74.0}$ & ${90.2}$ & ${81.2}$ & ${70.4}$ & ${80.7}$ & ${79.4}$\\      
      ~&HRFormer-Base~\cite{yuan2021hrformer}  & $256\times 192$& $43.2$M   & $14.1$G  &${75.6}$ & ${90.8}$ & ${82.8}$ & ${71.7}$ & ${82.6}$ & ${80.8}$\\
      ~&Swin-B$^\ddag$~\cite{liu2021Swin} & $256\times 192$& $94.0$M & $19.0$G & 73.7 &90.5&82.0&70.2&80.4&79.3\\
      ~&PVTv2-B2$^\ddag$~\cite{wang2022pvt} & $256\times 192$& $29.1$M & $5.1$G & 73.7& 90.5&81.2&70.0&80.6&79.1\\      
      ~&ViTPose-B~\cite{xu024vitpose+} & $256\times 192$& $90.0$M & $17.9$G  & ${75.8}$ & ${90.7}$ & ${83.2}$ & ${68.7}$ & ${78.4}$ & ${81.1}$ \\
      \cline{2-11}
      ~&HRFormer-Small~\cite{yuan2021hrformer}  & $384\times 288$& $7.7$M    & $7.3$G   & ${75.6}$ & ${90.3}$ & ${82.2}$ & ${71.6}$ & ${82.5}$ & ${80.7}$ \\
      ~&HRFormer-Base~\cite{yuan2021hrformer}  & $384\times 288$& $43.2$M   & $30.9$G  & ${77.2}$ & ${91.0}$ & ${83.6}$ & ${73.2}$ & ${84.2}$ & ${82.0}$ \\   
      \hline
      \multirow{6}{*}{\textit{\textbf{SSM}}}&Vim-S$^\ddag$~\cite{zhu2024vision}   & $256\times 192$& $28.0$M   & $6.1$G & 69.8 & 89.2 & 78.2 & 67.2 & 75.5 &76.0 \\
      ~&VMamba-T$^\ddag$~\cite{liu2024vmamba}   & $256\times 192$& $34.7$M   & $6.0$G   & ${74.4}$ & ${90.4}$ & ${82.3}$ & ${70.8}$ & ${81.0}$ & ${79.6}$ \\      
      ~&VMamba-B$^\ddag$~\cite{liu2024vmamba}   & $256\times 192$& $93.8$M   & $16.3$G   & ${74.8}$ & ${90.7}$ & ${82.1}$ & ${71.2}$ & ${81.5}$ & ${80.1}$ \\      
      ~& LocalVMamba-S$^\ddag$~\cite{huang2024localmamba} & $256\times 192$& $54.2$M   & $14.1$G & 74.1 & 90.4 & 81.8 & 70.9 & 80.4 & 79.9\\
      ~&MambaVision-B$^\ddag$~\cite{hatamizadeh2025mambavision}   & $256\times 192$& $102.9$M   & $24.6$G   & ${73.4}$ & ${90.1}$ & ${80.9}$ & ${69.7}$ & ${80.2}$ & ${78.9}$     \\
      ~&GroupMamba-B$^\ddag$~\cite{shaker2025groupmamba}   & $256\times 192$& $57.7$M   & $15.0$G  & ${73.2}$ & ${90.3}$ & ${81.1}$ & ${69.8}$ & ${79.8}$ & ${78.7}$ \\
      \hline
       \multirow{9}{*}{\textit{\textbf{SSM}}} 
       ~&HRVMamba-Tiny  & $256\times 192$& 2.3M   & 1.1G  & 69.5  & 88.3 & 77.0 & 66.2  &75.8  &76.1 \\
       ~&HRVMamba-Small  & $256\times 192$& $8.0$M   & $3.3$G  & ${74.6}$ & ${90.5}$ & ${81.7}$ & ${71.1}$ & ${81.0}$ & ${79.9}$ \\
       ~&HRVMamba-Base & $256\times 192$& $47.1$M   & $14.2$G & 76.5 & 90.9 & 83.6 & 73.0 &82.8 & 81.7\\
       \cline{2-11}
       \multirow{4}{*}{\textit{\textbf{(Ours)}}}&HRVMamba-Small$^{\dag}$ & $384\times 288$& $8.0$M    & $7.4$G & ${75.2}$ & ${90.3}$ & ${82.1}$ & ${71.7}$ & ${81.6}$ & ${80.3}$ \\
       &HRVMamba-Small & $384\times 288$& $8.0$M    & $7.4$G & ${76.4}$ & ${90.9}$ & ${83.3}$ & ${72.7}$ & ${83.0}$ & ${81.3}$ \\
       ~&HRVMamba-Base$^{\dag}$ & $384\times 288$& $47.1$M    & $32.0$G & 77.6 & 91.1 & \textbf{84.2} &\textbf{74.0} &84.2 & 82.4\\
       ~&HRVMamba-Base & $384\times 288$& $47.1$M    & $32.0$G & \textbf{77.7} & \textbf{91.2} & \textbf{84.2} &\textbf{74.0} &\textbf{84.3} & \textbf{82.5}\\
    \hline      
    \end{tabular}
  }
\end{table*}

\subsection{Dynamic Visual State Space (DVSS) Block}
We introduce the Dynamic Visual State Space (DVSS) Block. As illustrated in Fig.~\ref{fig:model} (b), the DVSS Block incorporates the Enhanced Spatial Inductive Bias Block (ESInB), the Deformable 2D-Selective-Scan (DSS2D) block, and a Feed-Forward Network (FFN) as its primary feature extraction blocks.

\textbf{Enhanced Spatial Inductive Bias Block (ESInB) Block.} 
While Visual Mamba establishes a global receptive field by treating images as token sequences and scanning them bidirectionally or along four directions, this sequential processing inherently disrupts the natural 2D spatial layout of visual data. As a result, the model lacks spatial inductive bias, which is crucial for capturing local textures, edges, and fine-grained details, particularly important in dense prediction tasks such as human pose estimation.

To compensate for this deficiency, we propose the Enhanced Spatial Inductive Bias Block, which explicitly incorporates local context through multi-scale convolutional operations. By leveraging multiple convolutional kernels of varying sizes, ESInB is designed to effectively model visual patterns at different spatial granularities, thereby restoring spatial awareness and enhancing the model’s capacity to generalize to various scales and poses.
Concretely, given an input feature map $\mathbf{X} \in \mathbb{R}^{H \times W \times C}$, we first partition it along the channel dimension into $G$ groups, where $G=4$ by default. Each group $\mathbf{X}_g$ is processed independently by a depthwise convolution with a distinct kernel size $K_g = 2g + 1$, allowing each group to focus on a specific spatial scale. This design introduces scale-specific locality into the model without significantly increasing computation.
The outputs of these convolutions, denoted as $\mathbf{Y}_g$, are then concatenated along the channel dimension and passed through a channel shuffle operation, which facilitates information exchange across different groups and prevents isolated feature learning. Finally, a GELU nonlinearity is applied to improve expressiveness:
\begin{align}
    [\vX_1, \vX_2, ..., \vX_G]&= \text{Split}(\vX, \text{axis=-1}),\label{eq:ESInB0}\\
    \vY_g &= \text{DWConv}_g(K_g\times K_g)(\vX_g), \label{eq:ESInB1}   
\end{align}
\begin{align}    
    \vY = \text{GELU}(\text{Shuffle}(\mathrm{Concat}([\vY_1, \vY_2, ..., \vY_G], \text{axis=-1})),
    \label{eq:ESInB2}
\end{align}
where $\text{Shuffle}$ denotes the channel shuffle operation. The ESInB block provides a lightweight yet effective enhancement mechanism for capturing local spatial dependencies, which are often diminished in pure state space architectures.

\textbf{Deformable 2D-Selective-Scan (DSS2D) Block.}
To further alleviate long-range forgetting and enhance spatial dependency modeling, we systematically explored a range of architectural design alternatives in Section~\ref{aba_ope}. Specifically, we investigated the use of dilated convolutions and deformable operations, both capable of enhancing long-range spatial interactions. The experimental results in Section~\ref{aba_ope}, along with our subsequent theoretical analysis, identify the deformable operation as a particularly effective mechanism.

This choice is theoretically grounded in the characteristics of SSMs. As described in prior work~\cite{shi2024multi}, an SSM applied to a sequence $\vx \in \bR^{1\times L \times C}$ defines the contribution of the $m$-th token to the $n$-th token ($m < n$) as:
\begin{align}
    \vC^{\top}_{n}\prod_{i=m}^{n}\tilde{\vA}_i \tilde{\vB}_m = \vC^{\top}_{n}\tilde{\vA}_{(m \rightarrow n)} \tilde{\vB}_m,
    \label{eq:s6_decay}
\end{align}
where $\tilde{\vA}_{(m \rightarrow n)} = \exp\left(\sum_{i=m}^n \Delta_i \vA\right)$ acts as an exponential decay factor along the sequence. Due to most $\Delta_i \vA$ being negative, $\tilde{\vA}_{(m \rightarrow n)}$ diminishes with larger $|n - m|$, leading to the long-range forgetting issue. This decay causes the model to underutilize distant contextual information and overemphasize low-level signals (see Fig.~\ref{fig_coco}, column 2).

To overcome this degradation, we propose the \textbf{DSS2D block}, which injects a deformable sampling operation~\cite{xiong2024efficient} into the State Space 2D-Selective-Scan (SS2D) structure. By allowing the model to dynamically sample from semantically important but spatially distant regions, DSS2D enhances the spatial expressiveness of SSMs and supports adaptive, context-aware feature integration.

Concretely, given a spatial input $\mathbf{X} \in \bR^{H \times W \times C}$, the deformable operation with $K = 9$ sampling points per group is defined as:
\begin{align}
    \vY_g &= \sum_{k=1}^K \mathbf{m}_{gk} \vX_g(p_o + p_k + \Delta p_{gk}), \\
    \vY &= \mathrm{Concat}([\vY_1, \vY_2, ..., \vY_G], \text{axis}=-1),
\end{align}
where $G$ is the number of groups, $p_k$ are regular grid offsets, $\Delta p_{gk}$ are learnable input-dependent offsets, and $\mathbf{m}_{gk}$ are modulation scalars. These allow the model to sample informative features beyond fixed neighborhoods.

This operation is embedded into the state update rule of the SSM, replacing the fixed token input $x(p_o)$ with a spatially aggregated representation:
\begin{align}
\vh_g(p_o) = \tilde{\vA} \vh_g(p_o-1) + \tilde{\vB} \sum_{k=1}^K \mathbf{m}_{gk} \vX_g(p_o + p_k + \Delta p_{gk}).
\end{align}
This integration mitigates the exponential decay of long-range dependencies by injecting and amplifying salient signals from relevant spatial regions into the SSM recurrence, enabling each hidden state update to incorporate flexible and task-adaptive spatial context, even from spatially distant regions.

\subsection{HRVMamba Architecture Instantiation}

We present the architectural configurations of the proposed HRVMamba model in Table~\ref{HRVMamba_arch}. The design follows a hierarchical multi-stage structure, where each stage progressively refines the spatial resolution and semantic abstraction. Specifically, in the $i$-th stage, we denote $B_i$ as the number of DSS2D blocks, $S_i$ as the expansion ratio of the state space block (SSM), $R_i$ as the MLP expansion ratio, and $M_i$ as the number of stacked blocks, which jointly control the model's representation capacity and computational complexity.

To accommodate different deployment scenarios and computational budgets, we instantiate HRVMamba in four variants: HRVMamba-Nano, HRVMamba-Tiny, HRVMamba-Small, and HRVMamba-Base. These versions differ primarily in terms of width, depth, and expansion ratios used in the SSM and MLP, enabling a flexible trade-off between accuracy and efficiency. The detailed configuration for each variant is summarized in Table~\ref{HRVMamba_variants}. 
We also integrate LPU block~\cite{guo2022cmt} and SE block~\cite{hu2018squeeze} for HRVMamba-Nano and HRVMamba-Tiny.
Our architectural characteristics are particularly beneficial for dense prediction tasks such as human pose estimation, where multi-scale spatial precision and long-range feature propagation are both crucial.

\begin{figure*}[t]
  \centering
  \includegraphics[width=0.9\textwidth]{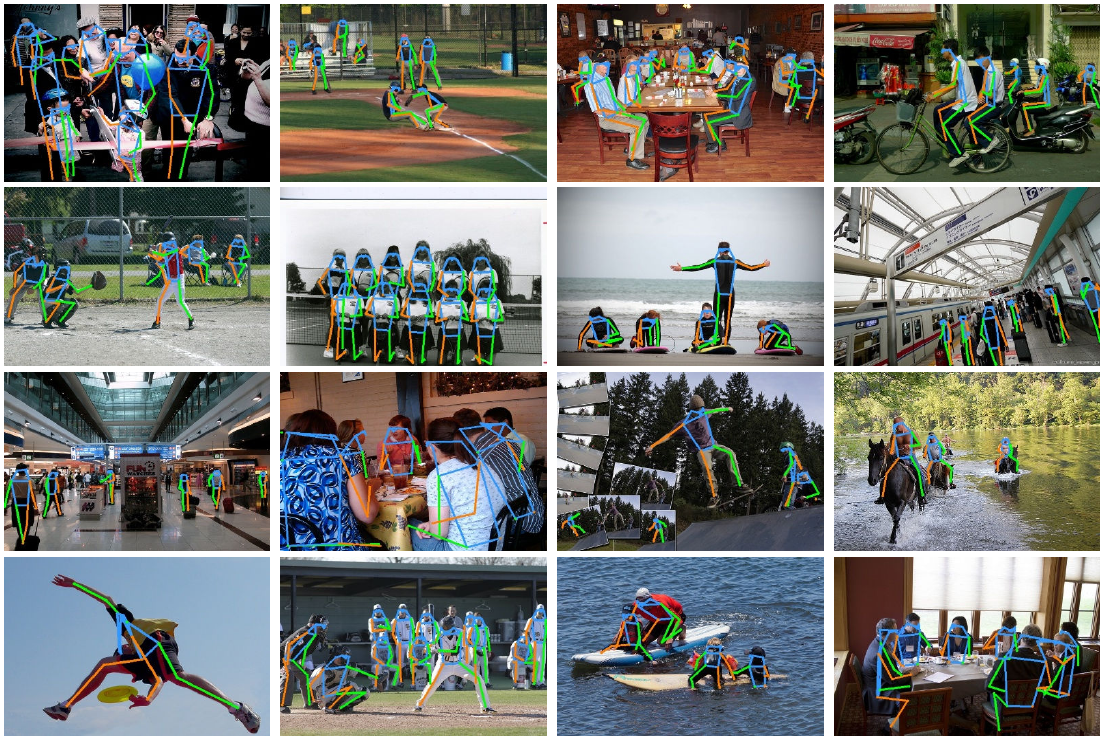}  
  \caption{\textbf{Example qualitative results on COCO pose estimation.} HRVMamba-Base is used as the backbone for pose estimation. The input size is $384\times288$.}
  \label{cocovis}
\end{figure*}
\begin{table*}[t]
  \centering
  \setlength{\tabcolsep}{4.0mm}
  \caption{\textbf{Comparison on the COCO pose estimation \texttt{test-dev} set.} $^\dag$ marks a model that is not pretrained, while $^\ddag$ signifies that the backbone uses the classic decoder from ViTPose. }
  \label{coco_pose_test}
  {
    \begin{tabular}{l|ccc|cccccc}
      \hline
      Method                                    & Input Size               & \#param.                 & FLOPs                   &
      $\operatorname{AP}$                       & $\operatorname{AP}^{50}$ & $\operatorname{AP}^{75}$ & $\operatorname{AP}^{M}$ & $\operatorname{AP}^{L}$ & $\operatorname{AR}$                                             \\
      \hline      
      HRNet-W$48$~\cite{wang2020deep}                  & $384\times 288$          & $63.6$M                  & $32.9$G                 & ${75.5}$                & ${92.5}$            & ${83.3}$ & ${71.9}$ & ${81.5}$ & ${80.5}$ \\
      PRTR~\cite{li2021pose}                    & $512\times 384$          & $57.2$M                  & $37.8$G                 & ${72.1}$                & ${90.4}$            & ${79.6}$ & ${68.1}$ & ${79.0}$ & ${79.4}$ \\
      TransPose-H-A$6$~\cite{yang2021transpose} & $256\times 192$          & $17.5$M                  & $21.8$G                 & ${75.0}$                & ${92.2}$            & ${82.3}$ & ${71.3}$ & ${81.1}$ & ${-}$    \\
      TokenPose-L/D$24$~\cite{li2021tokenpose}  & $384\times 288$          & $29.8$M                  & $22.1$G                 & ${75.9}$                & ${92.3}$            & ${83.4}$ & ${72.2}$ & ${82.1}$ & ${80.8}$ \\
      HRFormer-Small~\cite{yuan2021hrformer}    & $384\times 288$          & $7.7$M                   & $7.3$G                  & ${74.5}$                & ${92.3}$            & ${82.1}$ & ${70.7}$ & ${80.6}$ & ${79.8}$ \\
      HRFormer-Base~\cite{yuan2021hrformer}  & $384\times 288$          & $43.2$M                  & $30.9$G                 & ${76.2}$                & ${92.7}$            & ${83.8}$ & ${72.5}$ & ${82.3}$ & ${81.2}$ \\
      HRFormer-Base$^{\dag}$~\cite{yuan2021hrformer}  & $384\times 288$          & $43.2$M   & 30.9G & 76.0 & 92.6 & 83.6 & 72.9 &81.5 &81.0 \\
      Swin-L$^\ddag$~\cite{liu2021Swin} & $384\times 288$ & 207.9M &88.2G & 75.4 & 92.6 & 83.3& 72.0& 80.9&80.5 \\
      ViTPose-B~\cite{xu024vitpose+} &$256\times 192$& $90.0$M  & $17.9$G  & ${75.1}$    & ${92.5}$ & ${83.1}$ & ${72.0}$ & ${80.7}$ & ${80.3}$ \\
      VMamba-B$^\ddag$~\cite{liu2024vmamba} & $384\times 288$ & 93.8M & 36.6G & 75.3 & 92.7 &83.3&72.0&80.9& 80.3 \\
      \hline
      HRVMamba-Small & $384\times 288$& $8.0$M    & $7.4$G & 75.3 & 92.5 & 83.1 & 72.1 & 80.9 & 80.3\\
      HRVMamba-Base$^{\dag}$ & $384\times 288$& $47.1$M    & $32.0$G & 76.5 & 92.6 & 84.2 & \textbf{73.5} & 81.8 & 81.4\\
       HRVMamba-Base & $384\times 288$& $47.1$M    & $32.0$G & \textbf{76.7} & \textbf{92.8} & \textbf{84.4} & 73.4 & \textbf{82.2} & \textbf{81.5}\\
      \hline
    \end{tabular}
  }
\end{table*}

\section{Experiments}
\label{experiments}

\begin{table*}[t]
    \centering
    \caption{\textbf{Comparison with the state-of-the-art on ImageNet.} 
    ``iso.'', ''hie.'', ''hig.'' represent isotropic architecture without downsampling layers, hierarchical architecture, high-resolution architecture, respectively. $^\dag$ indicates that this implementation aligns the structure definition and classification header of our HRVMamba and adopts the basic block of HRFormer provided by MMPOSE~\cite{mmpose2020}.
    }
    \label{results_imagenet}  
    \setlength{\tabcolsep}{3.4mm}
        \begin{tabular}{c|l|l|c|c|r|r|c}
        \hline
 Type &Arch. & Model & Ref.& Input Size &\#Param (M) & FLOPs (G) & Top-1 Acc \\
 \hline
 \multirow{10}{*}{\textit{\textbf{iso.}}}& \multirow{2}{*}{\textit{\textbf{CNN}}} & ConvNeXt-S~\cite{liu2022convnet}& CVPR'2022  &  224$^2$ & 22 & 4.3 & 79.7 \\
  ~ &~ & ConvNeXt-B~\cite{liu2022convnet} & CVPR'2022 &  224$^2$ & 87 & 16.9 & 82.0 \\
 \cline{2-8}
  ~ &\multirow{2}{*}{\textit{\textbf{Trans.}}}  & DeiT-S~\cite{touvron2021training}&PMLR'2021 &224$^2$ & 22 & 4.6 & 79.8 \\
  ~ &~ & DeiT-B~\cite{touvron2021training}&PMLR'2021 & 224$^2$ & 87 & 17.6 & 81.8 \\
 \cline{2-8}
  ~ &\multirow{5}{*}{\textit{\textbf{SSM}}}  & S4ND-ViT-B~\cite{nguyen2022s4nd}&NeurIPS'2022 & 224$^2$ & 89 & - & 80.4 \\ 
  ~ & ~ & Vim-Ti~\cite{zhu2024vision} &ICML'2024& 224$^2$ & 7 & 1.1 & 76.9 \\
  ~ &~ & Vim-S~\cite{zhu2024vision} &ICML'2024& 224$^2$ & 26 & 4.3 & 80.5 \\ 
  ~ & ~ & VideoMamba-S~\cite{li2024videomamba} &ECCV'2024 & 
  448$^2$ & 26 & 16.9 & 83.3 \\
  ~ &~ &PlainMamba-L3~\cite{yang2024plainmamba} &BMVC'2024&  224$^2$ & 50 & 14.4 & 82.3 \\
  ~ & ~ & VideoMamba-M~\cite{li2024videomamba} &ECCV'2024& 
  576$^2$ & 75 & 83.1 & \textbf{84.0} \\   
  \hline
  \hline
 \multirow{9}{*}{\textit{\textbf{hie.}}}& \multirow{3}{*}{\textbf{\textit{CNN}}} & ConvNeXt-T~\cite{liu2022convnet}& CVPR'2022 & 224$^2$ & 29 & 4.5 & 82.1 \\
 ~ & ~ & ConvNeXt-B~\cite{liu2022convnet}& CVPR'2022 & 224$^2$ & 89 & 15.4 & 83.8 \\
 ~ & ~ &MambaOut-B~\cite{yu2025mambaout}& CVPR'2025&224$^2$ & 85 & 15.8 & 84.2 \\
 \cline{2-8}
  ~ &\multirow{2}{*}{\textit{\textbf{Trans.}}}  & Swin-T~\cite{liu2021Swin} & CVPR'2021& 224$^2$ & 28 & 4.5 & 81.3 \\
 ~ & ~ & Swin-B~\cite{liu2021Swin} & CVPR'2021&  224$^2$ & 88 & 15.4 & 83.5 \\
 \cline{2-8}
  ~ &\multirow{5}{*}{\textit{\textbf{\makecell[l]{SSM}}}}& VMamba-B~\cite{liu2024vmamba}&NeurIPS'2024 &  224$^2$ & 89 & 15.4 & 83.9 \\
  ~ &~ & LocalVMamba-S~\cite{huang2024localmamba}&ECCV'2024 &  224$^2$ & 50 & 11.4 &83.7 \\
   ~ &~&MambaVision-B~\cite{hatamizadeh2025mambavision}& CVPR'2025&  224$^2$ & 50 & 15.0 & 84.2 \\
   ~ &~&GroupMamba-B~\cite{shaker2025groupmamba}& CVPR'2025 &  224$^2$ & 57 & 14.0 & \textbf{84.5}\\
 \hline
 \hline
\multirow{9}{*}{\textit{\textbf{hig.}}}&  \multirow{4}{*}{\textit{\textbf{Trans.}}}& HRFormer-Nano$^\dag$~\cite{yuan2021hrformer,mmpose2020} &NeurIPS'2021&  256$^2$ & 12 & 1.9 & 74.3 \\
~ & ~ &HRFormer-Tiny$^\dag$~\cite{yuan2021hrformer,mmpose2020}&NeurIPS'2021 &  256$^2$ & 14  & 2.8 & 77.8 \\
~ & ~ & HRFormer-Small$^\dag$~\cite{yuan2021hrformer,mmpose2020} &NeurIPS'2021&  256$^2$ & 20 & 6.1 & 80.8 \\
 ~ & ~ & HRFormer-Base$^\dag$~\cite{yuan2021hrformer,mmpose2020} &NeurIPS'2021& 224$^2$ &57 & 14.5 & 83.3 \\
 ~ & ~ & HRFormer-Base~\cite{yuan2021hrformer} &NeurIPS'2021& 224$^2$ &50 & 13.7 & 82.8 \\
 \cline{2-8}
  ~ &\multirow{4}{*}{\textit{\textbf{SSM}}} 
  & HRVMamba-Nano & our& 256$^2$ & 12 & 1.9 & 74.8 \\
  ~ &~ & HRVMamba-Tiny & our&  256$^2$ & 14 & 2.8  & 78.6 \\
  ~ &~ & HRVMamba-Small & our&  256$^2$ & 20 & 5.8 & 81.3 \\
  ~ &~ & HRVMamba-Base& our &  224$^2$ & 61 & 15.8 & \textbf{84.2}  \\
        \hline
        \end{tabular}   
\end{table*}

\begin{figure}[t]
\centering
\includegraphics[width=0.42\textwidth]{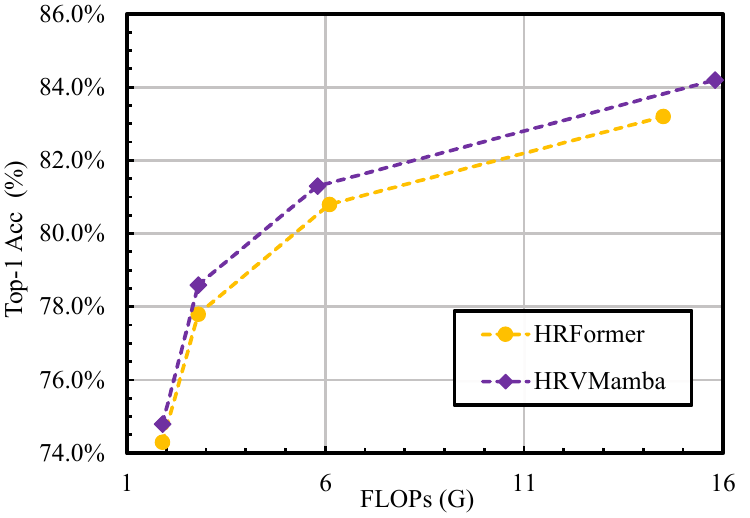}\\
\caption{\textbf{The trade-off between Top-1 accuracy and FLOPs on ImageNet
val set for high-resolution models.}
}
\label{fig_imagenet}
\end{figure}

\begin{table}[t]
\setlength{\tabcolsep}{1.8mm}
\centering
\caption{\textbf{Performance comparison for semantic segmentation.} We report the
mIoUs on ADE20K $\texttt{val}$. 'SS' and 'MS' denote evaluations performed at single-scale and multi-scale levels, respectively. All FLOPs are calculated with the input resolution fixed at $512 \times 2048$. Results of other methods are taken directly from their original papers. “-” indicates that the corresponding result was not provided in the original paper. }
\begin{tabular}{c|c|c|cc}
\hline
\multirow{2}{*}{Method} & \multirow{2}{*}{\#Param.} & \multirow{2}{*}{FLOPs} & \multicolumn{2}{c}{ADE20K} \\
\cline{4-5}
 &  && mIoU (SS) & mIoU (MS) \\
\hline
Swin-B~\cite{liu2021Swin}  & 121M & 1188G& 48.1 & 49.7 \\
ConvNeXt-B~\cite{liu2022convnet} & 122M & 1170G  & 49.1 & 49.9 \\ 
HRFormer-Base~\cite{yuan2021hrformer} & 56M  & 1120G  & 48.7 & 50.0 \\
VMamba-B~\cite{liu2024vmamba} & 122M  & 1170G & 51.0 & 51.6 \\
LocalVMamba-S~\cite{huang2024localmamba} & 81M  & 1095G  & 50.0 & 51.0 \\
MambaOut-B~\cite{yu2025mambaout} & 112M  & 1178G & 49.6 & 51.0 \\
MambaVision-B~\cite{hatamizadeh2025mambavision} & 126M  & 1342G & 49.1 & -  \\
\textbf{HRVMamba-Base} & 99M  & 1184G & \textbf{51.4} & \textbf{52.2} \\
\hline
\end{tabular}
\label{tabseg}
\end{table}

\begin{table*}[t]
  \centering
  \setlength{\tabcolsep}{2.8mm}
  \caption{\textbf{Performance comparison on the COCO pose estimation \texttt{val} set.} All models are trained from scratch without ImageNet pretraining.}
  \label{coco_aba_def}
  {
    \begin{tabular}{l|c|c|cccccc}
      \hline
      Model                                    & Input Size               &    Operation Strategy                &
      $\operatorname{AP}$                       & $\operatorname{AP}^{50}$ & $\operatorname{AP}^{75}$ & $\operatorname{AP}^{M}$ & $\operatorname{AP}^{L}$ & $\operatorname{AR}$                                             \\
      \hline  
      HRVMamba-Small & $256\times 192$& $3\times 3$ dilation convolution; dilation factors:1  & 71.5 & 89.1 & 78.7 & 68.4 & 77.5 & 77.0\\
      HRVMamba-Small & $256\times 192$&$3\times 3$ dilation convolution; dilation factors:3 & 70.5 & 89.0 & 78.1 & 67.3 & 76.6 & 76.0\\
      HRVMamba-Small & $256\times 192$&$3\times 3$ dilation convolution; dilation factors:5 & 70.1 & 88.5 & 77.9 & 67.0 & 76.3 & 75.8\\
      HRVMamba-Small & $256\times 192$&$3\times 3$ dilation convolution; dilation factors:1,3,5 & 71.2 & 88.9 & 79.1 & 68.1 & 77.4 & 76.7\\
      HRVMamba-Small & $256\times 192$& deformable operation & \textbf{73.1} & \textbf{89.6}  & \textbf{80.8}  &\textbf{69.9}  & \textbf{79.2} & \textbf{78.4}\\
      \hline
    \end{tabular}
  }
\end{table*}

\begin{table*}[t]
\centering
\setlength{\tabcolsep}{8.5mm}
\caption{\textbf{Ablation Experiments Results on COCO \texttt{val} set.} All models are not pretrained on the ImageNet. The HRVMamba-Small in Table~\ref{HRVMamba_variants} is the basic architecture setting. The input size is $256\times 192$. $^\dag$ denotes their is only $3\times 3$ depthwise convolution in ESInB Block.
}
    \begin{tabular}{c|c|c|c|cc}
    \hline        
       SS2D Block  & DSS2D Block & ESInB Block & ESInB in FFN  & $\operatorname{AP}$ & $\operatorname{AR}$ \\
    \hline
   \ding{52} & \ding{55} &\ding{55} & \ding{55}  & 70.3 & 75.9 \\ 
   \ding{55} & \ding{52} &\ding{55} & \ding{55}  & 72.7 & 78.0  \\ 
   \ding{55} & \ding{52} &\ding{55} & \ding{52} & 72.9 &  78.3 \\
   \ding{55} & \ding{52} &\ding{52} & \ding{55} & \textbf{73.1} & \textbf{78.4}  \\
   \ding{55}  & \ding{52} &\ding{52}$^\dag$ & \ding{55} & 72.8 & 78.3  \\
    \hline
    \end{tabular}
\label{aba}
\end{table*}

In this section, we compare the performance of HRVMamba with other state-of-the-art networks across several tasks: human pose estimation (COCO), image classification (ImageNet-1K), and semantic segmentation (ADE20K). We then analyze the ablation effects of the Multi-Resolution Parallel Architecture, the Deformable 2D-Selective-Scan block, and the Enhanced Spatial Inductive Bias Block.

\subsection{Human Pose Estimation}
\textbf{Training setting.} We evaluate HRVMamba on COCO dataset~\cite{lin2014microsoft} for human pose estimation, which comprises over 200,000 images and 250,000 labeled person instances with 17 keypoints. Our experiments are trained on the COCO \texttt{train} 2017 dataset, which includes 57,000 images and 150,000 person instances. The performance of our model is assessed on the \texttt{val} 2017 and \texttt{test-dev} 2017 sets, comprising 5,000 and 20,000 images, respectively.
For training and evaluation, we follow the implementation of MMPOSE~\cite{mmpose2020}. The AdamW optimizer is used, configured with a learning rate of 5e-4, betas of (0.9, 0.999), and a weight decay of 0.01.

\textbf{Results.} 
Table~\ref{coco_pose_val} presents the results on the COCO \texttt{val} dataset. HRVMamba consistently outperforms other CNN models, ViT models, and recent state-of-the-art SSM methods. With an input size of $256\times 192$, HRVMamba-Small achieves 74.6 $\operatorname{AP}$, exceeding FlashInternImage-B (74.1 $\operatorname{AP}$) while using only one-fifth of the FLOPs. HRVMamba-Base achieves 76.5 $\operatorname{AP}$, surpassing state-of-the-art SSM methods like Vim-S, VMamba-B, MambaVision-B, and GroupMamba-B. At similar computational complexity, HRVMamba-Base improves by 3.3 $\operatorname{AP}$ and 3.0 $\operatorname{AR}$ over GroupMamba-B. Additionally, HRVMamba-Base outperforms ViTPose-B by 0.7 $\operatorname{AP}$ and 0.6 $\operatorname{AR}$, with 50\% fewer parameters and 20\% fewer FLOPs.
With an input size of $384 \times 288$, HRVMamba-Small achieves a 0.8 $\operatorname{AP}$ improvement over HRFormer-Small. HRVMamba-Base achieves a 0.5 $\operatorname{AP}$ gain and a 0.5 $\operatorname{AR}$ improvement over HRFormer-Base with pretraining on ImageNet, while it gains 0.6 $\operatorname{AP}$ and 0.6 $\operatorname{AR}$ over HRFormer-Base without pretraining on ImageNet.
We present the pose estimation results of HRVMamba-Base pretrained on ImageNet in Fig.~\ref{cocovis}, which demonstrate that HRVMamba-Base effectively handles challenges such as viewpoint change, occlusion, and multiple persons.

We also provide comparisons on the COCO \texttt{test-dev} set in Table~\ref{coco_pose_test}. Our HRVMamba-Small achieves an $\operatorname{AP}$ of 75.3, outperforming ViTPose-B by 0.2 while using only one-eleventh of its parameters. It matches the performance of VMamba-B, but with just one-fifth of the FLOPs. Furthermore, HRVMamba-Base surpasses HRFormer-Base by 0.5 in $\operatorname{AP}$ and 0.4 in $\operatorname{AR}$ without pretraining on ImageNet. After pretraining on ImageNet, HRVMamba-Base achieves 76.7 $\operatorname{AP}$ and 81.5 $\operatorname{AR}$, setting a new state-of-the-art performance.

\subsection{ImageNet Classification}
\textbf{Training setting.} 
We conduct experiments on the ImageNet-1K dataset~\cite{deng2009imagenet}, which consists of 1.28M training images and 50K validation images across 1000 categories. HRVMamba is trained using the Swin Transformer~\cite{liu2021Swin} training framework on 80GB A100 GPUs. Specifically, we adopt the AdamW optimizer for 300 epochs with a cosine learning rate decay schedule and a 20-epoch linear warm-up phase. The initial learning rate is set to 0.002, and the weight decay is 0.05. We also apply exponential moving average (EMA), with MixUp and CutMix augmentation strategies set to 0.8 and 1.0, respectively. Due to inconsistencies between the official HRFormer implementation~\cite{yuan2021hrformer} and the HRFormer blocks employed in MMPOSE~\cite{mmpose2020}, we adopt the MMPOSE version, which is more commonly used in dense prediction tasks. As MMPOSE does not provide a classification head for HRFormer, we utilize a unified classification head based on HRVMamba for all comparisons. For HRFormer variants that are not defined in MMPOSE, such as HRFormer-Nano and HRFormer-Tiny, we follow the architectural specifications $(C_0, C_1, C_2, C_3, C_4)$ and $(B_1, B_2, B_3, B_4)$ of HRVMamba as detailed in Table~\ref{HRVMamba_variants}. To ensure a fair comparison, all HRFormer results reported in this section are based on our reimplementation under identical training settings as HRVMamba. The pretrained weights and source code are publicly released via our GitHub repository.

\textbf{Results.} 
Table~\ref{results_imagenet} compares HRVMamba with several representative CNN, ViT, and SSM methods. HRVMamba consistently demonstrates competitive performance across isotropic architectures~\cite{touvron2021training,li2024videomamba}, hierarchical architectures~\cite{liu2021Swin,liu2024vmamba}, and high-resolution architectures~\cite{yuan2021hrformer}. Specifically, HRVMamba-Base achieves a Top-1 accuracy of 84.2\%, while requiring only 19\% of the FLOPs consumed by VideoMamba-M, which reaches a comparable 84.0\% Top-1 accuracy. Notably, HRVMamba-Base achieves this result without relying on advanced training strategies, such as the LAMB optimizer used in MambaVision-B~\cite{hatamizadeh2025mambavision} (84.2\% Top-1 accuracy) or the distillation approach adopted in GroupMamba-B~\cite{shaker2025groupmamba} (84.5\% Top-1 accuracy). Although HRVMamba-Base slightly underperforms GroupMamba-B in image classification task under similar FLOPs, it substantially outperforms it in dense prediction tasks; as shown in Table~\ref{coco_pose_val}, HRVMamba-Base achieves 76.5 AP, compared to 73.2 AP by GroupMamba-B. 
Among high-resolution models, HRVMamba establishes a new state-of-the-art, with HRVMamba-Tiny, HRVMamba-Small, and HRVMamba-Base surpassing MMPOSE~\cite{mmpose2020}-reimplemented HRFormer-Nano, HRFormer-Tiny, HRFormer-Small, and HRFormer-Base by 0.5, 0.8, 0.5, and 0.9 points, respectively, and exceeding the original HRFormer-Base~\cite{yuan2021hrformer} by 1.4 points, all under comparable FLOPs.
Furthermore, as illustrated in Figure~\ref{fig_imagenet}, HRVMamba consistently delivers higher accuracy than HRFormer under equivalent computational budgets, underscoring its superior efficiency and strong potential for deployment on resource-constrained platforms.

\subsection{Semantic Segmentation}
\textbf{Training setting.} 
We adopt UPerNet~\cite{xiao2018unified} as the decoder head, with backbone networks initialized from ImageNet-1K pretrained weights~\cite{deng2009imagenet}. The ADE20K dataset serves as our primary benchmark, offering 150 fine-grained semantic classes across 20K training, 2K validation, and 3K test images.
Model optimization is performed using the AdamW optimizer with a base learning rate of $6 \times 10^{-5}$ and a weight decay of $5 \times 10^{-4}$. Training is conducted over 160k iterations with a batch size of 16. A linear learning rate warm-up is applied during the first 1,500 iterations. Input images are resized to a fixed resolution of $512 \times 512$.
Data augmentation strategies are consistent with standard practices, including random horizontal flips, scale jittering within a [0.5, 2.0] range, and photometric distortions. We report both single-scale and multi-scale performance on the ADE20K validation set for comprehensive evaluation.

\textbf{Results.} 
Table~\ref{tabseg} presents the results on the ADE20K \texttt{val} set. 
Under single-scale testing, HRVMamba-Base achieves a mIoU of \textbf{51.4}, surpassing recent SSM-based backbones such as MambaVision-B~\cite{hatamizadeh2025mambavision} by 2.3 mIoU, VMamba-B~\cite{liu2024vmamba} by 0.4 mIoU, and the CNN-based MambaOut-B~\cite{yu2025mambaout} by 1.8 mIoU.
With multi-scale testing, HRVMamba-Base further improves to \textbf{52.2} mIoU, outperforming HRFormer~\cite{yuan2021hrformer} by 2.2 mIoU and VMamba-B~\cite{liu2024vmamba} by 0.6 mIoU.
Overall, HRVMamba-Base consistently delivers higher segmentation accuracy under comparable computational budgets (FLOPs), demonstrating its strong capability for dense prediction tasks and highlighting its potential as an efficient and effective backbone for semantic segmentation.

\subsection{Ablation Experiments}

\subsubsection{Multi-resolution Parallel architecture} 
The results in Table~\ref{coco_pose_val} demonstrate that HRVMamba, utilizing the Multi-resolution Parallel architecture, achieved state-of-the-art performance in pose estimation. In particular, comparisons with other state-of-the-art SSM models such as VMamba~\cite{liu2024vmamba}, VMamba-B~\cite{liu2024vmamba}, MambaVision-B~\cite{hatamizadeh2025mambavision}, and GroupMamba-B~\cite{shaker2025groupmamba} highlight the advantages of the Multi-resolution Parallel architecture for dense prediction tasks.

\subsubsection{Operation Strategies to Alleviate Long-Range Forgetting} 
\label{aba_ope}
To investigate the impact of spatial operator design on mitigating long-range forgetting, we evaluate several operation strategies within the HRVMamba-Small framework. As shown in Table~\ref{coco_aba_def}, we explore dilated convolutions with different dilation factors and deformable operations, aiming to enhance long-range spatial interactions through adaptive spatial sampling.

We observe that using a standard dilation factor of 1 provides a solid baseline (AP: 71.5), while increasing the dilation factor to 3 or 5 leads to slight performance drops (AP: 70.5 and 70.1, respectively). This suggests that excessive dilation may disrupt local continuity and limit the model’s ability to maintain consistent spatial relationships across joints. Employing multiple dilation factors (1, 3, 5) slightly recovers the performance (AP: 71.2).
Remarkably, introducing deformable operations yields the best performance (AP: 73.1, AR: 78.4), highlighting their effectiveness in adaptively capturing spatially distributed dependencies. The learnable offsets in deformable convolutions enable the model to dynamically focus on semantically relevant positions, thereby strengthening the spatial coherence between distant keypoints and mitigating long-range forgetting.

These findings demonstrate that enhancing long-range spatial communication (deformable operation) rather than merely increasing receptive field size (dilated convolutions with different dilation factors) is crucial for robust pose estimation in visual SSMs. Deformable operations serve as a key mechanism to reinforce such dependencies in a data-driven, context-aware manner.

\subsubsection{Deformable 2D-Selective-Scan Block} 
As shown in Table~\ref{aba}, the DSS2D block improves $\operatorname{AP}$ by 2.4 points compared to the SS2D block (row 2 vs. row 1), demonstrating that incorporating deformable operations enhances Mamba's spatial feature extraction. Specifically, as shown in Fig.~\ref{fig_coco}, DSS2D focuses on high-level features related to the query patch in the early stage (S2), while SS2D targets low-level edge features. In the later stage (S3), DSS2D highlights human-related details, whereas SS2D tends to capture irrelevant background information. We think deformable operations enhance the features of high-level semantic relations between patches, allowing them to influence each other despite long-range decay (long-range forgetting issue).

\subsubsection{Enhanced Spatial Inductive Bias Block} 
HRFormer introduces depthwise convolution in the FFN to enhance the model's inductive bias. However, our experimental results in Table~\ref{aba} demonstrate that directly embedding the ESInB Block into the FFN (row 3 vs. row 2) yields only marginal improvement, suggesting limited effectiveness in this setting. In contrast, using the ESInB Block as a standalone block (row 4) leads to a notable performance boost, achieving the best AP of 73.1 and AR of 78.4. Furthermore, replacing the multi-scale convolutional kernels in the ESInB Block with a simple $3 \times 3$ depthwise convolution (row 5) reduces performance (AP: 72.8), highlighting the importance of multi-scale convolutions in enhancing local spatial representations and inductive bias.

\subsubsection{ImageNet Pretraining}
Table~\ref{coco_pose_val} shows that pretraining on ImageNet notably improves the performance of the smaller model, HRVMamba-Small, in pose estimation. However, for the larger variant, HRVMamba-Base, the gain is only marginal. This observation suggests that the pretraining strategy suitable for SSM-based models may differ from those commonly employed for CNNs and Vision Transformers, likely due to the unique recurrence and information propagation mechanisms of SSMs. These findings highlight a promising direction for future research: developing tailored pretraining methods for HRVMamba to fully exploit its potential. Exploring alternative strategies may lead to further performance gains across various model scales.

\section{Conclusion}
\label{conclusion}
Visual Mamba faces notable challenges in dense prediction tasks, including weak spatial inductive bias, long-range forgetting due to hidden state decay, and degraded spatial precision from low-resolution outputs. To address these limitations, we propose the Dynamic Visual State Space (DVSS) block, which enhances spatial inductive bias via multi-scale convolutions and alleviates long-range forgetting through input-adaptive deformable operations. Building upon the multi-resolution parallel architecture, we develop HRVMamba, a high-resolution visual state space model that preserves fine-grained spatial representations and supports efficient multi-scale feature learning.
Importantly, HRVMamba achieves a superior trade-off between computational complexity and accuracy, making it well-suited for deployment in resource-constrained scenarios, such as mobile devices, where computational budgets are limited. Extensive experiments across human pose estimation, image classification, and semantic segmentation demonstrate that HRVMamba delivers competitive performance compared to state-of-the-art CNN-, ViT-, and SSM-based models.

\normalem 
\bibliographystyle{IEEEtran}
\bibliography{main}

\begin{IEEEbiography}
[{\includegraphics[width=1in,height=1.25in,clip,keepaspectratio]{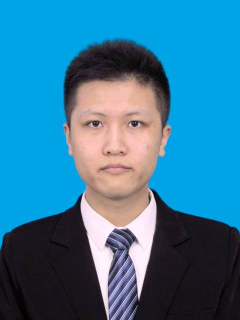}}]
{Hao Zhang} received a B.S. degree in information engineering from Xi’an Jiaotong University in 2021. He is currently pursuing a Ph.D. degree in artificial intelligence at Xi’an Jiaotong University. His research interests include neural network architecture design and Large Vision-Language Models.
\end{IEEEbiography}

\begin{IEEEbiography}
[{\includegraphics[width=1in,height=1.25in,clip,keepaspectratio]{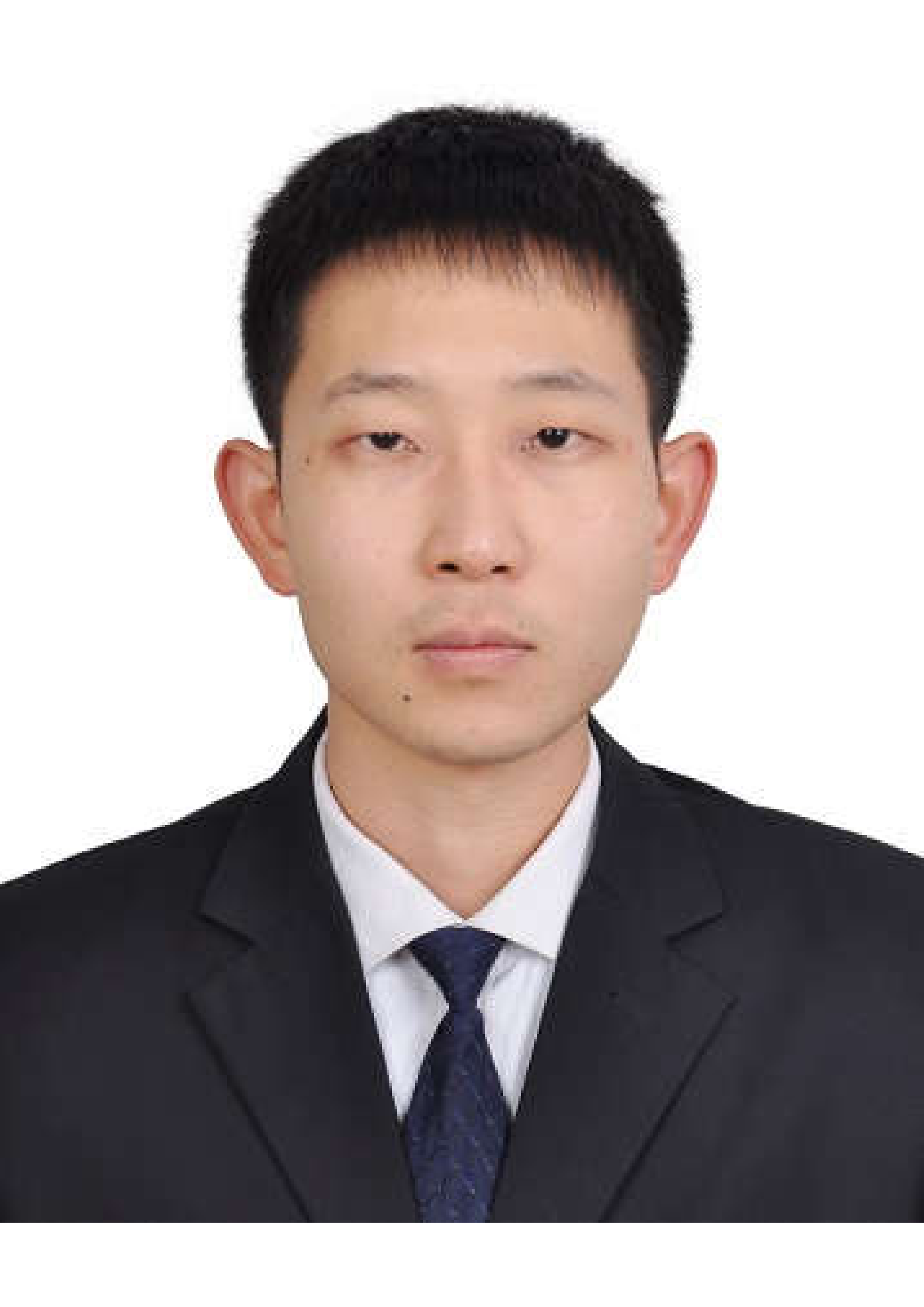}}]
{Yongqiang Ma} received an M.S. degree in software engineering from Xi'an Jiaotong University in 2015, and a Ph.D. degree in control science and engineering with Xi'an Jiaotong University in 2021. He is currently an assistant professor at Xi'an Jiaotong University. His research focuses on neuromorphic computing, spiking neural network, and cognitive Computing Model.
\end{IEEEbiography}

\begin{IEEEbiography}
[{\includegraphics[width=1in,height=1.25in,clip,keepaspectratio]{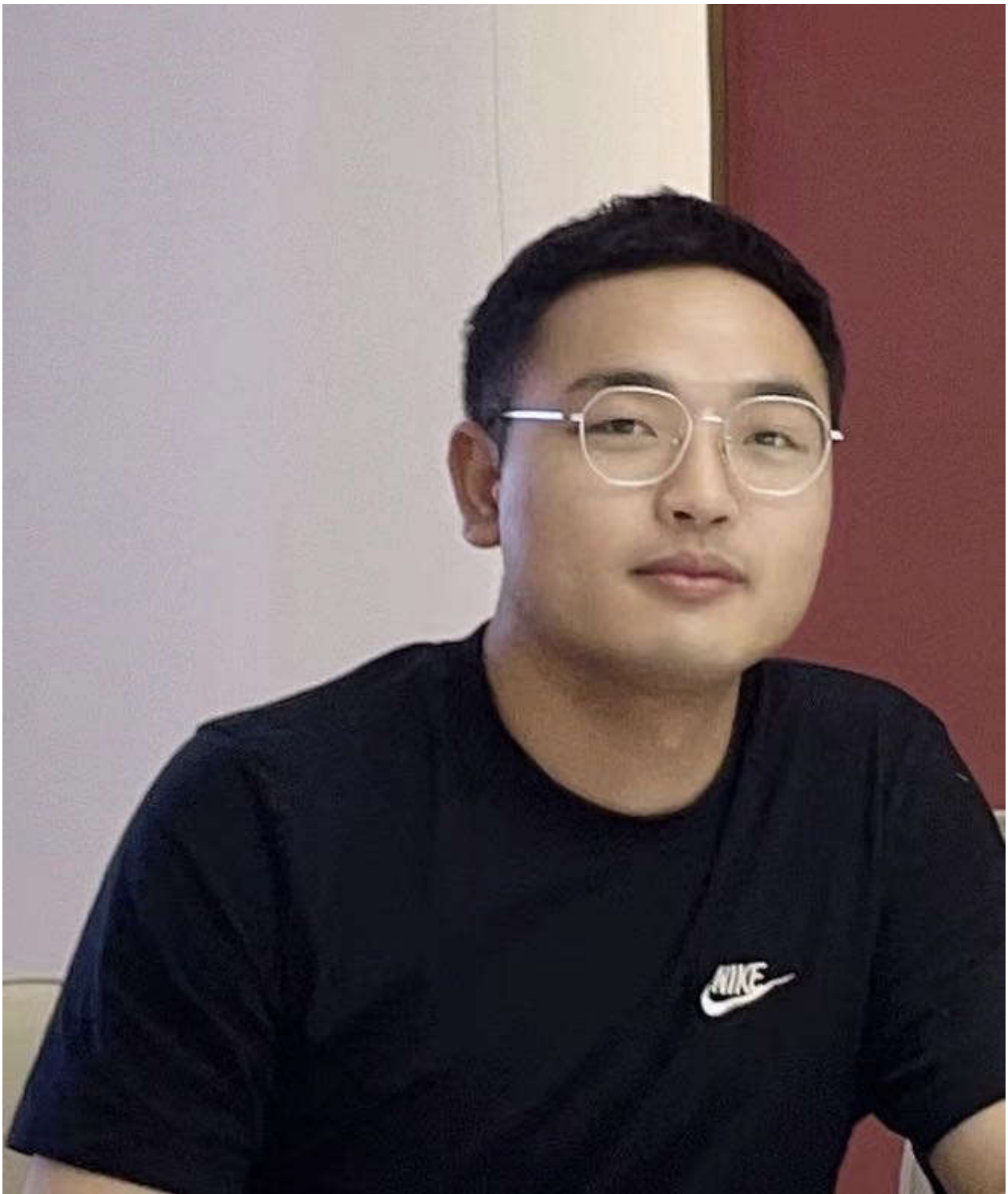}}]
{Wenqi Shao} received the Ph.D. degree from Multimedia Lab, the Chinese University of Hong Kong (CUHK) in 2022. 
Now he is a researcher at Shanghai Artificial Intelligence Lab, Shanghai, China.
His research interests lie in the pre-training, evaluation, applications of multimodal foundation models, as well as compression techniques and hardware codesign for large models.
\end{IEEEbiography}

\begin{IEEEbiography}
[{\includegraphics[width=1in,height=1.25in,clip,keepaspectratio]{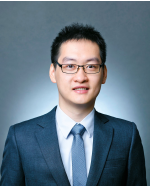}}]
{Ping Luo} received the Ph.D. degree in information engineering from the Chinese University of Hong Kong (CUHK).
He is currently an associate professor with the Department of Computer Science, University of Hong Kong (HKU). He was a postdoctoral fellow in CUHK
from 2014 to 2016. His research interests include machine learning and computer vision. He has published more than 100 peer-reviewed articles in top-tier conferences and journals.
\end{IEEEbiography}

\begin{IEEEbiography}[{\includegraphics[width=1in,height=1.25in,clip,keepaspectratio]{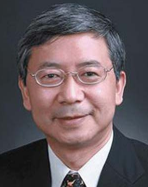}}]{Nanning Zheng} graduated from the Department of Electrical Engineering, Xi’an Jiaotong University, Xi’an, China, in 1975, and received the M.S. degree in information and control engineering from Xi’an Jiaotong University in 1981 and the Ph.D. degree in electrical engineering from Keio University, Yokohama, Japan, in 1985. He joined Xi’an Jiaotong University in 1975, where he is currently a professor and the director of the Institute of Artificial Intelligence and Robotics. His research interests include computer vision, pattern recognition, and machine learning. Dr. Zheng became a member of the Chinese Academy of Engineering in 1999. He is the Chinese Representative on the Governing Board of the International Association for Pattern Recognition. 
\end{IEEEbiography}

\begin{IEEEbiography}[{\includegraphics[width=1in,height=1.25in,clip,keepaspectratio]{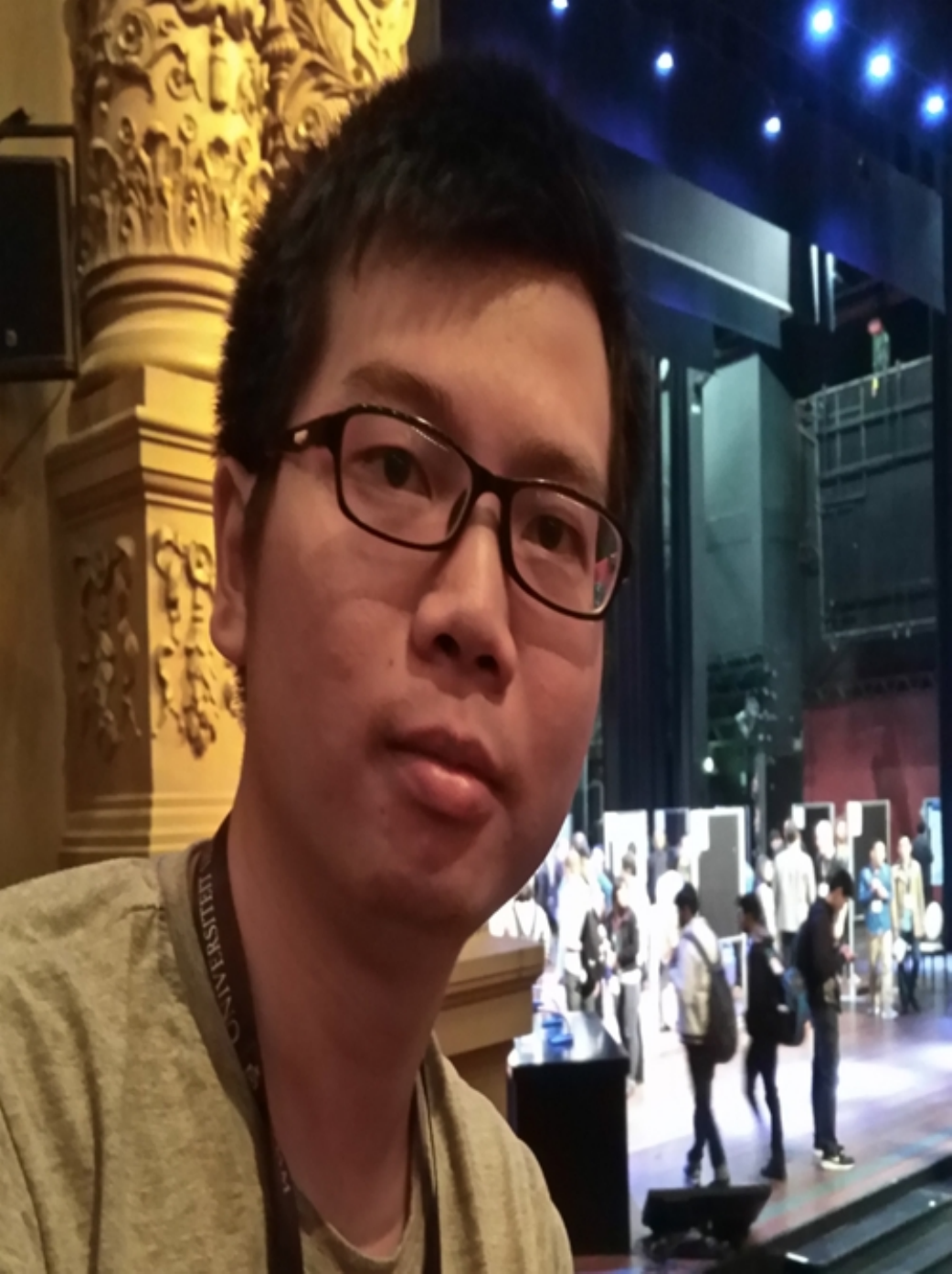}}]{Kaipeng Zhang}
received an M.S. degree from National Taiwan University, Taipei, Taiwan in 2018, and a Ph.D. degree from the University of Tokyo, Tokyo, Japan in 2022.
Now he is a researcher at Shanghai Artificial Intelligence Lab, Shanghai, China. His current research interests include face analysis, active learning, and foundation vision models.
\end{IEEEbiography}

\newpage

\vfill
\end{document}